\documentclass[letterpaper, 10 pt, conference]{ieeeconf}  

\IEEEoverridecommandlockouts                              

\usepackage{booktabs}
\usepackage{colortbl}
\usepackage{xcolor}
\usepackage{pifont}
\usepackage{graphicx}
\usepackage{subcaption}
\usepackage{wrapfig}
\usepackage{amsmath,amssymb,amsfonts}
\usepackage{url} 
\usepackage{multirow}
\definecolor{mygreen}{RGB}{114, 182, 161} 
\definecolor{myorange}{RGB}{233, 150, 117} 

\newcommand{\cmark}{\textcolor{green}{\ding{51}}}  
\newcommand{\xmark}{\textcolor{red}{\ding{55}}}    

\newcommand{\figvspace}{\vspace{-1.1em}} 

\title{\LARGE \bf
MoDeSuite: Robot Learning Task Suite for Benchmarking Mobile Manipulation with Deformable Objects
}

\author{Yuying Zhang$^{1}$, Kevin Sebastian Luck$^{2}$, Francesco Verdoja$^{1}$, Ville Kyrki$^{1}$, Joni Pajarinen$^{1}$ 
\thanks{$^{1}$ Department of Electrical Engineering and Automation, Aalto University, Espoo, Finland}
\thanks{$^{2}$ Faculty of Science, Vrije Universiteit Amsterdam, Amsterdam, Netherlands }
}

\begin{document}

\maketitle
\thispagestyle{empty}
\pagestyle{empty}

\begin{abstract}

Mobile manipulation is a critical capability for robots operating in diverse, real-world environments. However, manipulating deformable objects and materials remains a major challenge for existing robot learning algorithms. While various benchmarks have been proposed to evaluate manipulation strategies with rigid objects, there is still a notable lack of standardized benchmarks that address mobile manipulation tasks involving deformable objects.

To address this gap, we introduce MoDeSuite, the first Mobile Manipulation Deformable Object task suite, designed specifically for robot learning. MoDeSuite consists of eight distinct mobile manipulation tasks covering both elastic objects and deformable objects, each presenting a unique challenge inspired by real-world robot applications. Success in these tasks requires effective collaboration between the robot's base and manipulator, as well as the ability to exploit the deformability of the objects. To evaluate and demonstrate the use of the proposed benchmark, we train two state-of-the-art reinforcement learning algorithms and two imitation learning algorithms, highlighting the difficulties encountered and showing their performance in simulation. 
Furthermore, we demonstrate the practical relevance of the suite by deploying the trained policies directly into the real world with the Spot robot, showcasing the potential for sim-to-real transfer. We expect that MoDeSuite will open a novel research domain in mobile manipulation involving deformable objects.
Find more details, code, and videos at \url{https://sites.google.com/view/modesuite/home}.
\end{abstract}

\section{Introduction}

Mobile manipulation is a complex robotics challenge, integrating robot navigation and object manipulation. Mastering these abilities enables robots to perform intricate and dynamic tasks, ranging from fetching and placing~\cite{mm_diffusion} and opening doors~\cite{mm_opendoor} to fruit harvestings~\cite{mm_cherrypick} and human rescue~\cite{mm_rescue} in disaster scenarios. Many of these tasks involve manipulating deformable objects, a particularly challenging problem that requires further research into robust and adaptable robotic learning techniques~\cite{sandakalmm2022motion_review}.

Deformable objects introduce further unique challenges for mobile manipulators due to their shape variability, which directly impacts manipulation strategies. These challenges can be categorized based on deformation type: plastic or elastic~\cite{materials_book}. Plastic deformation results in permanent structural changes, requiring robust policies that account for irreversible modifications. In contrast, elastic deformation involves temporary, reversible changes, necessitating precise modeling and real-time control to avoid exceeding the material's elastic limit. Both types demand advanced perception, planning, and control strategies to ensure reliable task execution.

\begin{figure}
    \centering
    \includegraphics[width=0.45\textwidth]{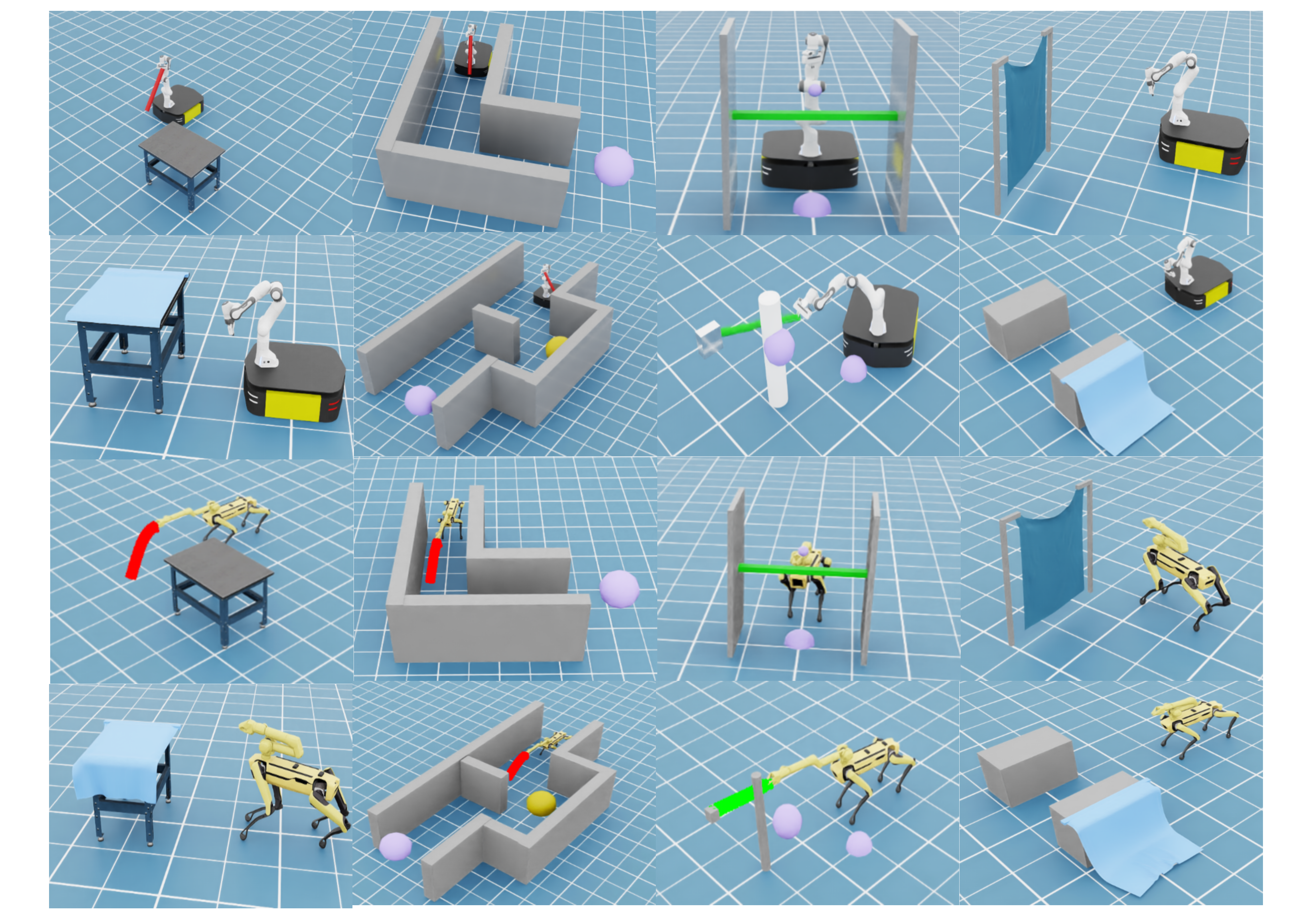}
    \caption{
    MoDeSuite features diverse tasks requiring coordinated navigation and manipulation of elastic and plastic deformable materials using both wheeled and legged robots in constrained environments.
    \figvspace{}
}
\label{fig_tasks}
\end{figure}
Despite advancements in both mobile manipulation and deformable object manipulation, there is currently no standardized benchmark that integrates these two domains. This gap limits the ability to systematically develop, evaluate, and compare algorithms across different methodologies. Existing benchmarks primarily focus on either rigid-body mobile~\cite{bench_manipulathor,zhu_robosuite} or static deformable object manipulation~\cite{bench_softgym,bench_reform}, leaving mobile deformable manipulation relatively underexplored. Furthermore, real-world experimentation is time-consuming, costly, and difficult to standardize, especially for data-driven approaches like reinforcement learning (RL) and imitation learning (IL). A well-designed simulation-based benchmark could significantly accelerate progress by providing a controlled, reproducible, and scalable testing environment.

To address this gap, we introduce MoDeSuite, a standardized task suite specifically designed for mobile deformable manipulation, consisting of eight diverse tasks. As shown in Fig.~\ref{fig_tasks}, MoDeSuite includes two types of mobile manipulators, three types of action spaces, two types of observation spaces, and support for both elastic and plastic object manipulation. Within each task, users can switch between different robots, observation modalities, and action spaces, facilitating flexible experimentation.

\begin{table*}
\centering
\caption{Comparison of different manipulation frameworks. The check(\cmark) denotes the presence of the feature. In Supported Dynamics, Plastic denotes plastic deformable objects, including clothes and rope, and Elastic denotes elastic objects such as rubber and foam. In Robotics Platforms, Mobile-M denotes a manipulator with a mobile base, and Legged-M denotes a manipulator with a legged mobile base.}
    \label{tab:table_compare}
    \small
    \setlength\tabcolsep{3pt}
    \begin{tabular}{c|c|c|*2c|*2c|c}
        \toprule
        \multicolumn{3}{c}{}
        &\multicolumn{2}{c}{Supported Dynamics} 
        &\multicolumn{2}{c}{Robotic Platforms}
        \\
        Category & Name             & Physics Engine   & Plastic & Elastic  & Mobile-M & Legged-M&Note  \\    
        \midrule            
       \multirow{6}{*}{Mobile} & Behavior-1k~\cite{bench_behavior}      & PhysX5          &\cmark &\xmark  & \cmark& \xmark&Daily Activities\\
        &AI2THOR~\cite{bench_ai2thor}     & Unity           & \xmark& \cmark    & \cmark&\xmark&Indoor Scene\\
        &TDW Tansport\textsuperscript{*}~\cite{bench_tdwtransp} & PhysX,Flex,Obi    & \cmark&  \cmark   &\cmark & \xmark&Transport Challenge\\
        &Habitat~\cite{bench_habitat3} & Bullet    & \xmark&  \xmark    &\cmark & \cmark&Indoor Scene\\
        &ManiSkill3~\cite{bench_maniskill3} & PhysX    & \xmark&  \xmark  &\cmark & \xmark&Limited Mobile Manipulation Tasks\\
        &ORBIT~\cite{bench_orbit} & PhysX    & \cmark&  \xmark   &\cmark & \cmark&FEM,Particle System\\

        \midrule            
       \multirow{7}{*}{Deformable} & 
        DeformableRavens~\cite{bench_DeformableRAvens} &   Bullet   &\cmark & \cmark   &  \xmark & \xmark& FEM\\
        &DEDO~\cite{bench_DEDObench}             &   Bullet       &\cmark&   \cmark       & \cmark &\xmark& Particle System\\
        &DAXBench~\cite{bench_daxbench}         &   DAX             &\cmark & \cmark         & \xmark &  \xmark  & Tabletop, Particle System\\
        &PlasticineLab~\cite{bench_plasticinelab}         & DiffTaiChi  & \cmark &   \cmark     &\xmark   & \xmark & End-effector, Particle System\\
        &Reform~\cite{bench_reform}        &  AGX   & \xmark & \cmark      & \xmark  & \xmark & End-effector, FEM \\
        &SoftGym~\cite{bench_softgym}       &  Flex   &\cmark & \xmark      &\xmark      &\xmark&Particle System\\
        &DexGarmentLab~\cite{bench_dexgarmentlab}       &  Physx   &\cmark & \cmark      &\xmark      &\xmark&Particle System, FEM\\
        \midrule
         \rowcolor[HTML]{AAACED}
         Both& MoDeSuite(ours)             &   Physx          & \cmark       & \cmark   & \cmark   &  \cmark& Particle System, FEM\\
        \bottomrule    
    \end{tabular}
    \figvspace{}
\end{table*}

MoDeSuite is developed within Isaac Lab~\cite{bench_orbit} and utilizes the high-fidelity simulator Isaac Sim~\cite{Isaacsim}, enabling efficient training through parallelized environments. Success in these tasks requires agents to exploit object deformability while simultaneously overcoming the dual challenges of navigation and manipulation. To support research in this area, MoDeSuite provides pre-configured models with camera sensors, leveraging the latest advancements in photorealistic rendering and high-fidelity physics simulation.

We benchmark four state-of-the-art learning algorithms, two from imitation learning and two from reinforcement learning, and provide a dataset for offline imitation training. To validate the practical significance of our proposed tasks, we implement similar environments in the real world, demonstrating that our benchmark can facilitating sim-to-real transfer.

We believe MoDeSuite represents a crucial step forward in the development of mobile deformable manipulation by providing a unified platform for research, benchmarking, and algorithm development. By bridging the gap between mobile manipulation and deformable object interaction, we aim to accelerate progress in both fields. The codebase and detailed installation instructions will be made publicly available upon paper acceptance.

\section{Related Work}
\subsubsection*{Mobile Manipulation with Deformables} Mobile manipulation involving deformable objects presents significant challenges due to the coordination required between the mobile platform, robotic arms, and the complex dynamics of deformable materials. Most existing approaches that address this problem rely on planning-based methods~\cite{mode_algo_plan,  mode_algo_plan_aksoy}, which often struggle to generalize across diverse tasks and environments. Recently, data-driven methods, particularly imitation learning (IL) and reinforcement learning (RL), have shown promise in mobile manipulation~\cite{ma_RLlegmm_2022, honerkamp_rlwmm_2021} and deformable tasks like shape control~\cite{agl_learning_shapecontrol} and cloth manipulation~\cite{ ILdeform_2019, ILdeform_demo_2022}, including mobile deformable scenarios~\cite{Demobot}. However, these methods require either diverse demonstrations or extensive interactions, highlighting the need for simulation environments that offer realistic and diverse tasks to support scalable learning.

\subsubsection*{Mobile Manipulation Benchmark} Several existing mobile manipulation benchmarks are designed for specific domains such as underwater~\cite{bench_underwater}, aerial~\cite{bench_aerial_2020, bench_aerial_gym, bench_drones}, assistive~\cite{bench_assistibegym}, or rover-based platforms~\cite{bench_rover}. More general task suites~\cite{bench_brs, bench_ai2thor, bench_mm_dual} aim to evaluate coordination between the mobile base and manipulator, primarily using rigid objects~\cite{zhu_robosuite,bench_habitat2,bench_habitat3} or with limited support for deformable manipulation~\cite{bench_behavior,bench_tdwtransp, bench_ai2thor}. Consequently, standardized environments for evaluating mobile manipulation involving deformable objects remain limited.

\subsubsection*{Deformable Manipulation Benchmark} Conversely, several benchmarks focus on deformable object manipulation, such as DAXBench~\cite{bench_daxbench}, DeformableRavens~\cite{bench_DeformableRAvens}, DEDO~\cite{bench_DEDObench}, and SoftGym~\cite{bench_softgym}, which primarily target tasks involving plastic deformable objects, such as rope~\cite{bench_shoelace} and cloth, often excluding elastic-specific scenarios. Reform~\cite{bench_reform} and PlasticineLab~\cite{bench_plasticinelab} incorporate both elastic and plastic deformables but are constrained by fee-based simulators or pre-programmed task setups. ORBIT-Surgical~\cite{bench_surg} focuses exclusively on surgical tasks. Furthermore, these benchmarks predominantly focus on stationary robotic arms (e.g., Sawyer, Franka~\cite{bench_softgym}, UR5~\cite{bench_DeformableRAvens}) and lack support for robot mobility. Although ORBIT~\cite{bench_orbit} includes mobile manipulators and deformable objects, it focuses on framework design and fails to address the unique integration of deformable manipulation and mobile manipulation. Table~\ref{tab:table_compare} provides a detailed comparison of these general benchmarks.

\begin{figure}
    \centering
    \includegraphics[width=\linewidth]{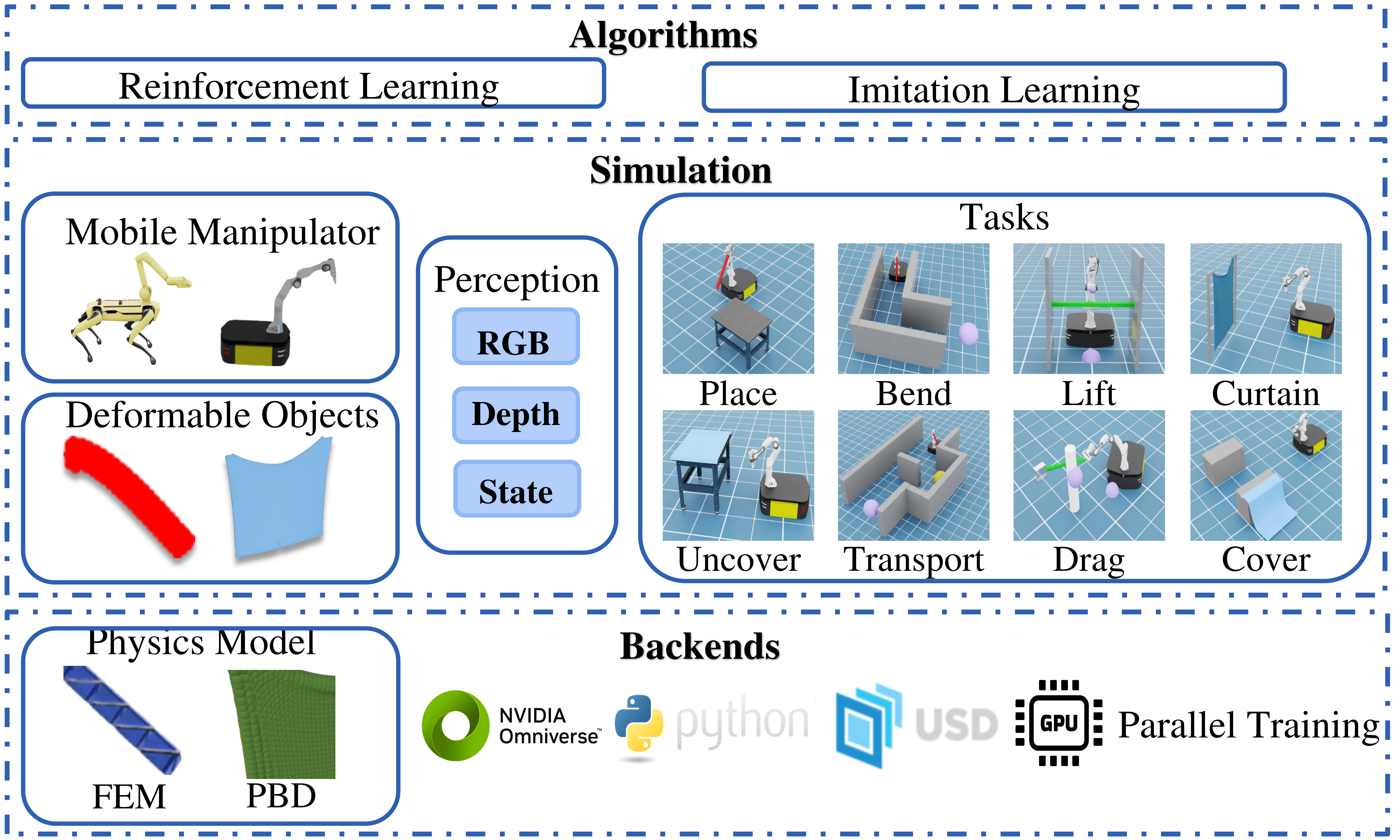} 
    \caption{Overview of MoDeSuite. MoDeSuite, built on the NVIDIA Omniverse, supports reinforcement and imitation learning in a simulated environment with two mobile manipulator types. It offers RGB, depth, and state-based perception inputs for eight deformable manipulation tasks featuring different shapes. The simulation uses FEM and PBD physics models and supports GPU-accelerated parallel training.
}
\label{fig_overview}
\figvspace{}
\end{figure}

\section{MoDeSuite}

In this paper, we introduce the Mobile Deformable Manipulation (MoDe) task suite, designed to accelerate algorithm development in robotic manipulation. As illustrated in Fig.~\ref{fig_overview}, MoDeSuite supports both reinforcement learning and imitation learning approaches within a simulated environment featuring mobile manipulators interacting with a variety of deformable objects. The suite comprises eight tasks, including five newly designed elastic manipulation tasks plus three plastic deformable tasks originally proposed in~\cite{Demobot}, where they were showcased as challenging tasks for imitation learning. Our primary contribution over~\cite{Demobot} is the development of a scalable and extensible benchmark that addresses several limitations of the original implementation, such as the lack of locomotion support and coordinated base control. MoDeSuite is composed of two main elements: (1) a simulation framework, and (2) a diverse set of tasks that serve as examples and baselines for customization and evaluation. The details of these components are described below.

\subsection{Framework Overview}
The framework includes a variety of different environments and two different robotic platforms. The robot perceives the environment through multimodal observations, including RGB and depth images, proprioceptive states, and object-specific information. Based on these inputs, it generates actions to interact with the environment. The interaction dynamics are powered by the NVIDIA PhysX engine, which provides accurate modeling of rigid and deformable body physics. This enables realistic simulation of complex contact interactions and collisions, which is critical for tasks involving deformable materials.

The environment is built using the Isaac Sim graphical interface. For deformable objects, MoDeSuite uses two distinct simulation methods to capture the dynamics of different types of object deformability. Elastic bodies are simulated using element finite methods(FEM) which use a combination of a finite number of tetrahedral meshes for modeling. The FEM has been used in linear elastic object simulation to simulate various deformation features efficiently and accurately. Meanwhile, the plastic deformable objects, such as curtains and tablecloths, are simulated with the position-based-dynamics (PBD) particle simulation systems to handle the large deformations without stability issues~\cite{modeling_review}.

\subsection{Mobile Manipulator and Action Space}

\subsubsection*{Mobile Manipulator} MoDeSuite incorporates two types of robotic settings, wheeled and legged manipulators, to accommodate a wide range of application scenarios. In the wheeled robot configuration, the Franka Panda robot arm~\cite{Franka} is mounted on the Ridgeback wheeled base~\cite{Ridgeback}, a midsize indoor robot platform from Clearpath Robotics. The legged configuration utilizes the Spot body~\cite{Spot} and its associated arm~\cite{Spot_Arm}, both from Boston Dynamics. We aim to propose tasks that require simultaneous control of the arm and mobile base. Therefore, the robot action consists of the arm and body action, $a_{\text{robot}}=(a_{\text{base}}, a_{\text{arm}})$. We detail the action settings according to different robots below. 

\subsubsection*{Action Space} For the wheeled mobile base configuration, the base action represents the joint velocities, denoted as $a_{\text{base}} =(v_x,v_y,w_z)$, where $v_x$ and $v_y$ are the linear velocities along the \( x \)- and \( y \)-axes, respectively, and $w_z$ represents the rotational velocity around the \( z \)-axes. For the manipulator, we support two types of control modes: (1) joint space control, where the action is specified as joint position \(a_{\text{arm}}\in \mathbb{R}^n\), where \( n \) is the number of joints and (2) end-effector pose control, where the action defines a desired pose \(a_{\text{arm}}\in SE(3)\), a 6D vector comprising translational and rotational components. 

For the quadrupedal manipulator, we support two types of controllers for both the Spot body and arm, resulting in four possible control configurations. The Spot base has 12 degrees of freedom and can be controlled either by a separate locomotion controller or by directly controlling the 12 joints. Thus, the base action is defined as either $a_{\text{base}} \in \mathbb{R}^n$, where \( n \) is the number of joints or $a_{\text{base}} =(p_x,p_y,r_z)$, where the $p_x$ and $p_y$ represent the linear translation along the \( x \)- and \( y \)-axes, respectively, and $r_z$ represents the rotation around the $z$-axis. Similarly to the Franka arm, the Spot arm action includes two types: (1) joint space, $a_{\text{arm}} \in \mathbb{R}^n$, and (2) end-effector pose, $a_{\text{arm}} \in SE(3)$. Consequently, the number of possible Spot action dimensions ranges from 10 to 18, depending on the selected control configuration.

\subsubsection*{Discrete action setting} To simplify controlling the agent and improve data collection efficiency, we also provide a discrete action space that can be mapped to the keyboard. Specifically, the robot has the following discrete actions implemented: (1) body move forward, (2) body move left, (3) body move right, (4) body move backward, (5) body turn left, (6) body turn right, (7) hand move forward, (8) hand move backward, (9) hand move left, (10) hand move right, (11) hand move up,  (12) hand move down, (13) hand grasping, and (14) hand release. 

\subsection{Observation Spaces}
Our task suite accommodates two types of observation spaces: image-based and state-based observations. The image-based observation is obtained from the RGB-D camera mounted on the robot. This observation format is straightforward to transfer to real-world robots; however, its high dimensionality introduces challenges during training. In contrast, the state-based observation offers detailed information about both the deformable objects and the robot's internal state, which reduces the training difficulty. However, acquiring such data in real-world settings is more challenging.

The observation for each task is divided into three primary components: the robot state, the deformable object state, and additional environmental information (e.g., obstacle information, the target positions). Thus, the general form of the observation for all tasks is defined as: $O=(s_{\text{r}}, s_{\text{o}}, s_{\text{e}})$, in which $s_{\text{r}}$ represents the robot state, $s_{\text{o}}$ is related to the deformable objects, and $s_{\text{e}}$ for the remain task-related information, such as the target position and the obstacle position.

\subsubsection*{Robot State} The robot state, $s_{\text{r}}$, is noted as $s_r=(p_r,q_r,q,\dot{q})$, where $p_r\in \mathbb{R}^2$ and  $q_r\in \mathbb{R}^4$ represent the 2D position and orientation (as a quaternion) of the mobile robot platform, respectively. \( q\in \mathbb{R}^n\) denotes the \( n \) joint positions of the manipulator, while \(\dot{q}\in \mathbb{R}^n\) corresponds to the joint velocities.

\subsubsection*{Object State} For elastic objects, the state is represented by the positions of the simulation elements, which are defined as: $s_o = \{e_i \in \mathbb{R}^3 \}_{i=1}^N$, where \( N \) is the number of FEM elements, and each \( e_i \) denotes the 3D position of the \( i \)-th element. For plastic deformable objects, such as cloth-like materials, we use particle-based simulations. The state is represented by the positions of particles. This is defined as: 
$s_o = \{p_i \in \mathbb{R}^3 \}_{i=1}^M$, where \( M \) is the number of tracked particles, and each \( p_i \) denotes the 3D position of the \( i \)-th particle.

\subsubsection*{Additional Information} The $s_e$ component includes additional task-relevant information, including target and obstacle positions. Specifically, it is defined as: $s_e=(g_r,g_o,p_o)$ with $g_r,g_o,p_o\in \mathbb{R}^3$, where $g_r$ denotes the target position for the robot, $g_o$ denotes the target position for the objects, and $p_o$ indicates the position of the obstacles. The target positions are visualized in purple in Fig~\ref{fig_tasks}.

\subsubsection*{Image-based Observation}  In the case of image-based observation, the robot receives RGB-D data from its camera. Rather than using the raw image data directly, we define the observation as $O=s_{\text{env}}=\phi(I)$, where $I$ denotes the RGB image, and $\phi$ represents the image encoder. For this task suite, we employ the DiNOv2 image encoder~\cite{DINOv2} to process the RGB input.

\begin{figure}
    \centering
    \includegraphics[width=0.48\textwidth,]{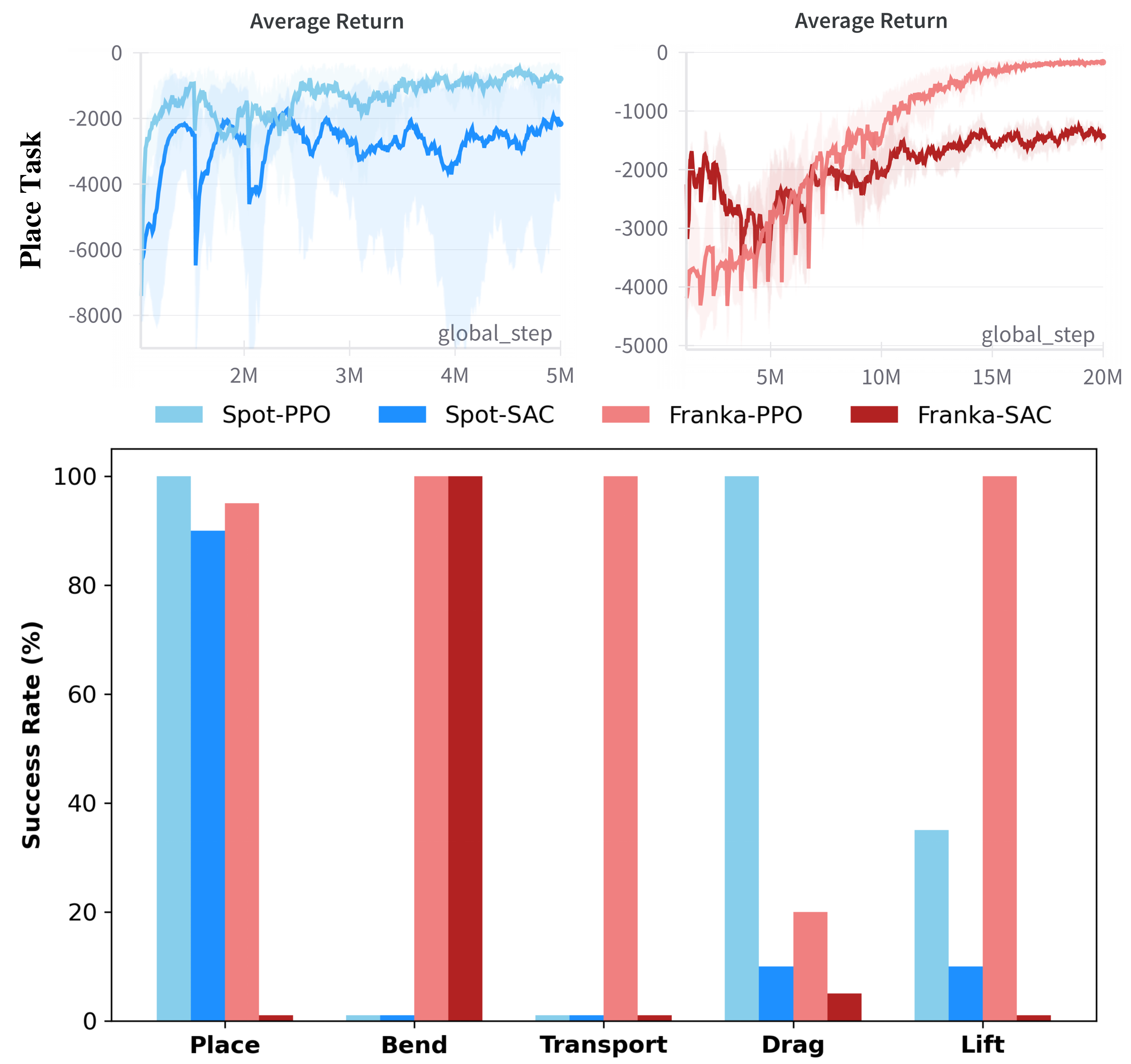} %
    \caption{The performance of SAC and PPO algorithms on MoDeSuite tasks (Place, Bend, Transport, Lift, Drag) is evaluated using state-based observations and two robot platforms (Franka and Spot). Curves represent the mean return over 5 seeds, with shaded areas showing standard deviation for the Place task. Bar plots display success rates over 20 trials. Results highlight the impact of robot morphology and algorithm choice on task effectiveness.
}
\label{fig_rlplot}
\figvspace{}
\end{figure}

\subsection{MoDeSuite: Task Suite}
\label{sec:action_space}
Inspired by scenarios commonly encountered in daily life, MoDeSuite offers five elastic deformation tasks---Place, Bend, Transport, Drag, and Lift---and three plastic deformation tasks---Cover, Uncover, and Curtain---to support research in deformable mobile manipulation. Figure~\ref{fig_tasks} illustrates the five tasks implemented in the simulation environment. Below, we provide a detailed description of each task along with the corresponding evaluation metrics.

\subsubsection*{Place} This task requires the robot to position the elastic rod onto the table located beyond the reach of its manipulator, requiring the simultaneous control of both the mobile base and arm. The robot needs to efficiently exploit both the rod's deformability and its mobility to successfully complete the task. The reward function is defined as the negative sum of three components: (i) the distance from the rod’s endpoint to the table, (ii) the distance from the robot to the table, and (iii) a stability penalty (which is zero for wheeled configurations). This formulation encourages the robot to minimize both distances and maintain balance throughout the motion. The task is considered successful if the endpoint of the rod is positioned on the table. 

\subsubsection*{Bend} This task challenges the robot with mobile manipulation involving a complex load, specifically a long elastic rod. In this task, the robot must move through an L-shaped corridor while holding the elastic rod. The flexibility of the rod allows the robot to navigate through confined spaces by bending it appropriately.  The reward function comprises three components: (i) the distance from the rod’s endpoint to the target, (ii) the distance from the robot to the target, and (iii) a stability penalty. The success of this task is measured by the distance between the rod's endpoint and the purple target located at the entrance of the corridor. 

\subsubsection*{Transport} This task extends the Bend task, requiring the mobile manipulator to navigate toward a target position while navigating around a large obstacle placed in the middle of the path. This obstacle significantly increases the complexity of both path planning and rod manipulation in a confined environment. In addition to spatial constraints, a major challenge is avoiding locally optimal behaviors, such as the robot becoming trapped between corners and failing to make progress toward the final goal. To mitigate this, we introduce an intermediate target that encourages the robot to successfully navigate past the first corner. The reward consists of three components: (i) the distance from the rod’s endpoint to the middle and final target, (ii)  the distance between the robot and the two targets, and (iii) a stability penalty. Task success is determined by the distance between the rod’s endpoint and the final target.

\subsubsection*{Drag} This task challenges the robot to manipulate an elastic belt that is fixed at one end of a cube, while an obstacle blocks the path between the robot and the target. The robot needs to lift and stretch the belt over the obstacle and place it on the other side, all while maintaining its body near the designated body target. During execution, the belt undergoes significant stretching, increasing the force between the robot’s gripper and the elastic material. This added tension introduces instability in the robot’s control, particularly for legged robots, making the collaboration between the mobile base and arm manipulation critical. To encourage the robot to utilize its mobility rather than relying solely on arm movement, an additional body target is introduced. The reward consists of three components: (i) the distance between the belt’s midpoint and the belt target, (ii) the distance between the robot’s body and the body target, and (iii) a stability penalty. The task is considered solved if the robot is close to the body target and moves the belt to the other side of the obstacle close to the belt target.

\subsubsection*{Lift} This task is inspired by real-world scenarios such as operating a roller shutter or manipulating other elastic objects that require vertical movement. An elastic belt is suspended between two high walls, with both endpoints fixed to the walls. The robot must first lift the belt to create sufficient clearance before navigating through the corridor to reach the final target position. This task is particularly challenging because the robot must approach the belt, lift it high enough to pass underneath, and then pass through the opening while maintaining stability. The friction between the elastic belts and the end-effectors, as well as the robot’s movement after lifting, further complicates the task. Effective execution requires precise force application and coordinated motion to prevent the belt from obstructing the robot’s path and the instability of the locomotion. The reward consists of three components:(i) the distance between the belt’s midpoint and the belt target, (ii) the distance between the robot’s body and the body target, and (iii) a stability penalty. The success metric for this task is the distance between the belt’s midpoint and the target position, and the distance between the robot and the robot's target. 

\subsubsection*{Uncover}  In this task, the robot must approach the table and remove the table cover by pulling it in a specific direction, ensuring the cloth folds properly during removal. The table is large enough that successful execution requires coordinated movement of both the robot’s body and arm. A key challenge is that grasping is essential but causes only minimal movement, resulting in subtle visual changes that make it hard for the agent to perceive progress. Additionally, the robot must carefully manipulate the cloth to prevent unintended entanglements or collisions. The task is evaluated using a binary sparse reward. The agent receives a reward if the table cover is completely removed and its handle has been pulled beyond the other side of the table. To solve this task, the robot must grasp the cloth, pull it away from the table, and avoid any collisions with the table.

\subsubsection*{Cover} This task requires the robot to grasp a fabric and use it to cover the gap between two objects. This is a long-horizon task that involves multiple steps: the robot must first approach the deformable fabric, grasp it, and then move it to fully cover the designated gap between two cubes. The gap’s covering necessitates coordinated movement of both the robot’s body and arm.  As in the previous task, the grasping action also presents a challenge due to the minimal visual change it produces. Additionally, the presence of the fabric can obstruct the robot’s movement, particularly for legged robots, leading to partial observations and increased task complexity. A binary sparse reward function is used to evaluate success. The agent receives a reward only when the gap is covered by the fabric and the fabric’s handle has been moved to another cube.

\subsubsection*{Curtain} This task requires the robot to approach a hanging curtain, use its arm to move the curtain aside, and then navigate its body through the opening without any collisions. This task introduces multiple challenges, including partial observability and potential failures in the inverse kinematic solver, which can prevent successful execution. Additionally, the curtain may slip from the robot’s end effector, further increasing task difficulty. A binary sparse reward function is used to evaluate success. The robot receives a reward only if it successfully moves past the curtain without any collisions. 

\begin{figure}
    \centering
    \includegraphics[width=0.48\textwidth]{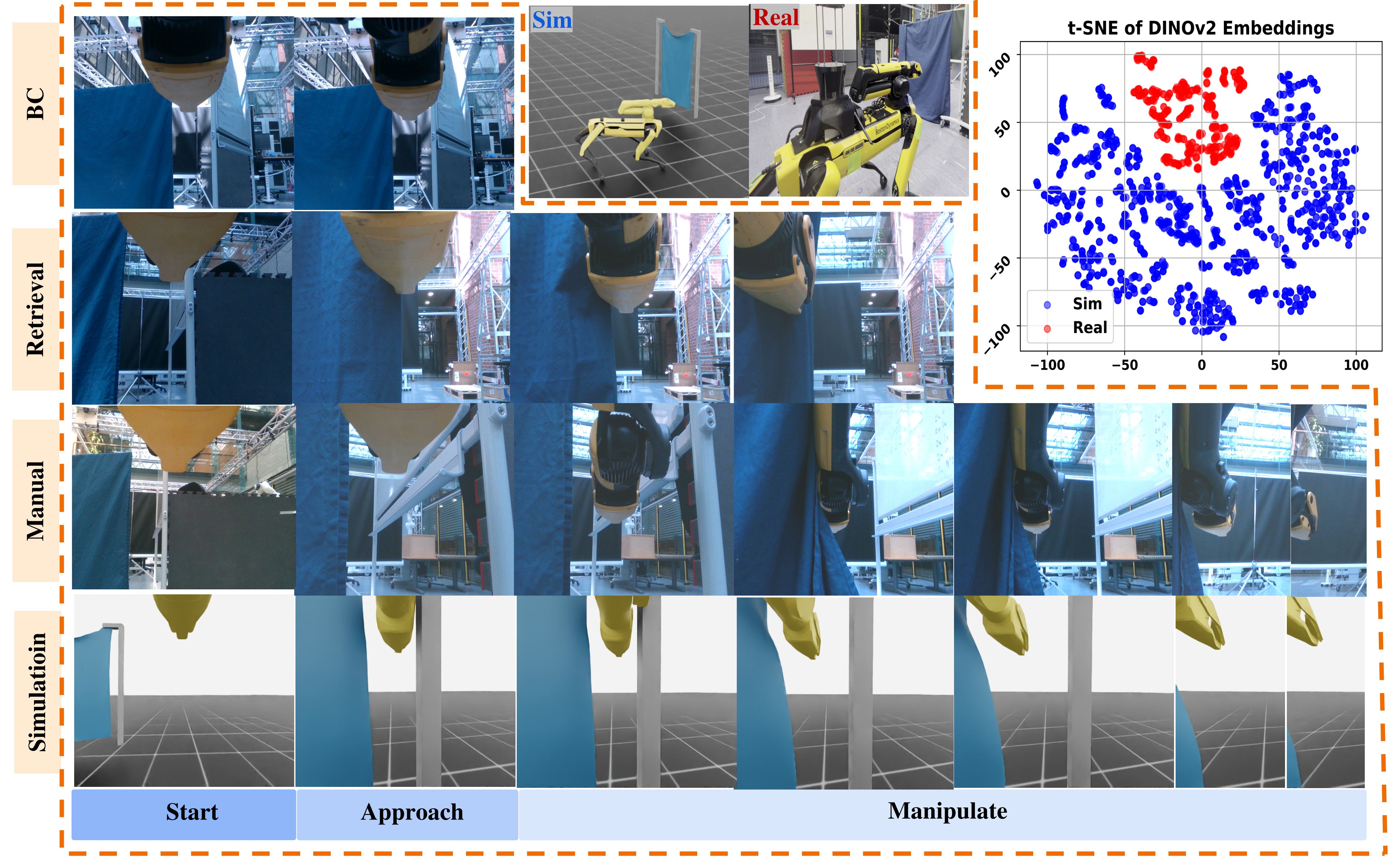} 
    \caption{The Curtain task's Sim-to-Real visual comparison features image sequences for various control strategies: behavior cloning (BC), image retrieval guidance (Retrieval), manual teleoperation, and simulation. The top-right panel presents a t-SNE plot of DiNOv2 image embeddings, highlighting the visual domain gap between simulation (blue) and real data (red). The top-middle section shows the physical setups in simulation and the real world.
}
\label{fig_real_curtain}
\figvspace{}
\end{figure}

\section{Experiments}

In our experiments, we aim to systematically evaluate the effectiveness of our task suite in training agents capable of performing mobile manipulation tasks involving deformable objects. Specifically, we study: (1) the ability of agents to learn from interaction and demonstration in simulation, (2)the impact of different input state-based versus image-based perception on learning, and (3) the zero-shot transferability of learned policies from simulation to the real world without fine-tuning. Our experimental design incorporates a variety of observation types (state and image), learning paradigms (reinforcement learning and imitation learning), and evaluation metrics to assess both learning efficacy and sim-to-real performance gaps. Further details for each setting are provided below.

\begin{table}
\centering
\caption{Sim-to-real performance comparison for SAC and PPO on Place and Drag tasks. Steps represent control actions needed per task, with real-world operation at 10 Hz and simulation at 60 Hz. SR indicates success rate, based on 10 real-world trials and 20 simulation trials.}
\label{tab:sim2real}
\begin{tabular}{lcccccc}
\toprule
\multicolumn{2}{c}{} & \multicolumn{2}{c}{\textbf{Place}} & \multicolumn{2}{c}{\textbf{Drag}}\\
\cmidrule(lr){3-4}  \cmidrule(lr){5-6}
\textbf{} & \textbf{Method} & \textbf{SR(\%)} & \textbf{Steps} & \textbf{SR(\%)} & \textbf{Steps} \\
\midrule
\multirow{2}{*}{Sim} 
& SAC & 90 & 217.6 & 10 & 92.5 \\
& PPO & 100 & 83.4 & 100 & 81.3 \\
\midrule
\multirow{2}{*}{Real} 
& SAC & 90 & 172.1 & 0 & -- \\
& PPO & 100 & 62.6 & 100 & 32.9 \\
\bottomrule
\end{tabular}
\figvspace{}
\end{table}

\subsection{Reinforcement Learning} 
We evaluate Proximal Policy Optimization (PPO)~\cite{schulman2017ppo} and Soft Actor-Critic (SAC)~\cite{haarnoja2018sac} algorithms on the elastic tasks with both mobile manipulator settings. Both methods are implemented using the high-performance framework RL Games~\cite{rl_games}. These experiments aim to assess the challenges of manipulating elastic objects in mobile settings.

We use state-based observations as input, which include the positions of four points uniformly distributed along the linear elastic objects. Therefore, the observation is $O=(s_r, s_o, s_e)$, where $s_o = \{e_i \in \mathbb{R}^3\}_{i=1}^{4} $ represents the state of the elastic object. The action space includes control for both the mobile base and the manipulator arm. In the wheeled mobile manipulator setting, control is applied in the joint action space. In the legged setting, we employ a pretrained locomotion controller for the base and apply joint space control to the manipulator arm, as described in Section~\ref{sec:action_space}.

We report the training results from five independent runs for each algorithm with different random seeds. In Fig.~\ref{fig_rlplot}, we present the training curves for the place task, showing the average episode return and the standard deviation across random seeds. Bar plots located in the upper-right corner show the success rate of trained agents over 20 evaluation trials. 

\begin{figure}
    \centering
    \includegraphics[width=0.48\textwidth]{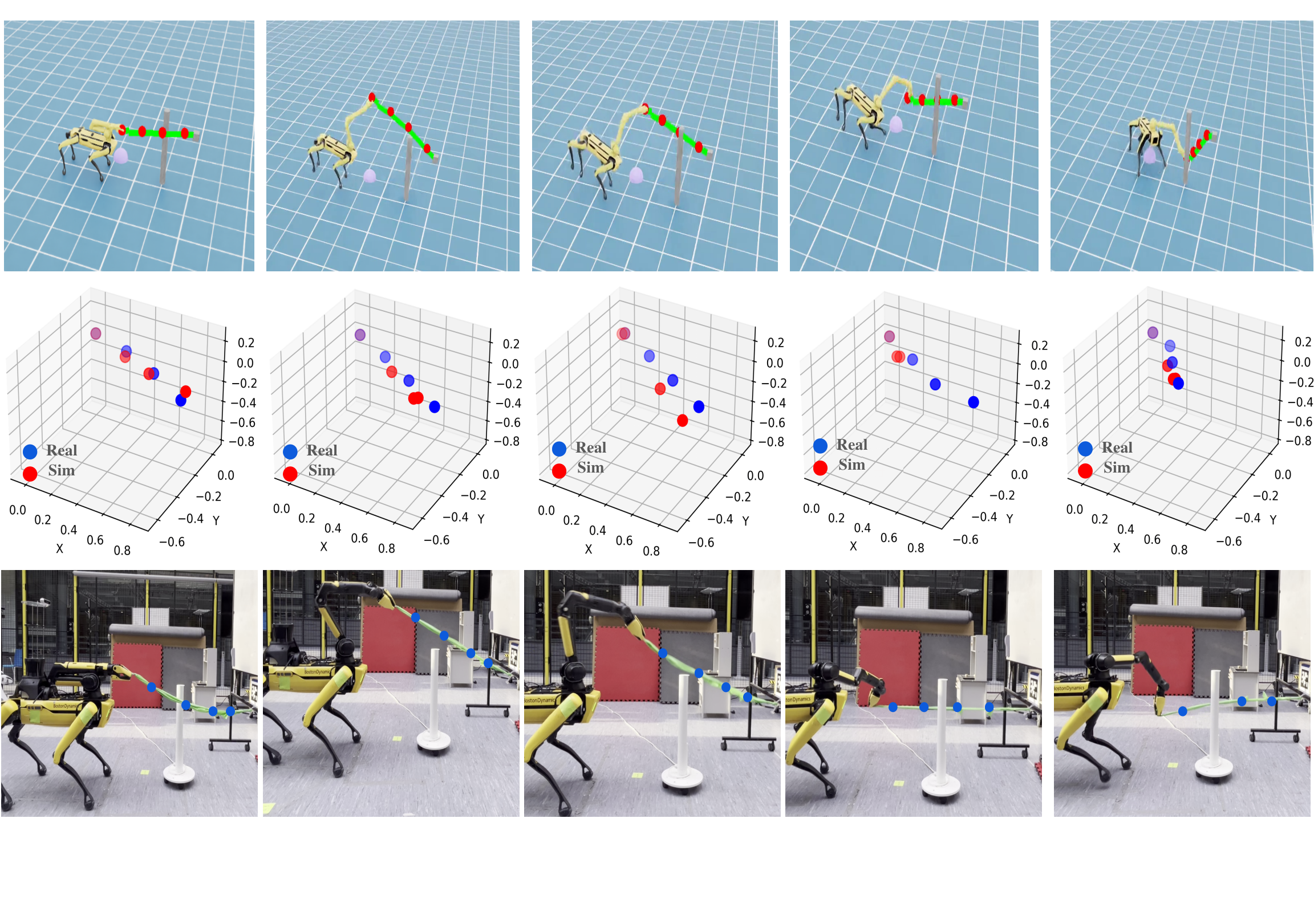} 
    \vspace{-2.5em}
    \caption{Sim-Real comparison for the Drag task shows trajectories from a PPO agent. The top row depicts the robot's rollout in simulation. The middle row compares state trajectories of the deformable object in simulation (red points) against real-world (blue points), highlighting similarities and differences. The bottom row showcases the agent's performance in a physical setting, illustrating real-world dynamics.
}
\label{fig_drag_real}
\figvspace{}
\end{figure}

The results across tasks reveal key insights into both algorithmic performance and the impact of robot morphology. Overall, tasks in the legged robot setting (Spot) are notably more challenging than those in the fixed-base manipulator setting (Franka), primarily due to the added complexity of maintaining balance during manipulation. This is especially evident in confined-space tasks such as Bend and Transport, where Spot must coordinate whole-body movements under more restrictive conditions.

From an algorithmic perspective, PPO consistently outperforms SAC across nearly all tasks and robot configurations, with the exception of the Bend task in the legged (Spot) setting. In this specific scenario, the Spot robot frequently collides with the narrow walls or the elastic objects and loses balance, leading to task failure. We hypothesize that improved reward shaping and more careful hyperparameter tuning could enhance performance in such constrained environments. Overall, the results indicate that all tasks in the MoDeSuite are solvable, yet remain challenging for current state-of-the-art reinforcement learning algorithms, highlighting the need for further advances in both algorithm robustness and interaction modeling.

\subsection{Imitation Learning} 
We implement two imitation learning algorithms on the legged deformable tasks with purely image-based observation: one is a classical supervised behavior cloning (BC) algorithm~\cite{BC}, the other is a simple retrieval-based method (Retrieval)~\cite{Demobot} with only state similarity. Both algorithms are trained using feature extraction from RGB data via a visual foundation model~\cite{DINOv2}, paired with expert actions. Thus, the observation space is defined as $ob=\phi(I_{\text{rgb}})$, where $I_{\text{rgb}}$ is the RGB image. The agents are trained on a dataset consisting of 30 demonstrations per task, which were collected using a keyboard controller. 

Table~\ref{tab:il_result} presents the success rate, evaluated over 20 rollouts, in an environment identical to the data-collection environment. Despite the identical settings, the agents still encounter several failures due to the challenges posed by the mobile manipulator and the dynamics of the deformable objects. Beyond the typical challenges associated with imitation learning, the accuracy limitations of the locomotion controller contribute to some failure cases.

\begin{table}
\centering
\caption{Success rates of different methods across three Deformable Mobile Manipulation tasks. Both models are trained using the full dataset of 30 demonstrations and evaluated over 20 trials per task.
}
\label{tab:il_result}
\begin{tabular}{c*{3}{c}}
\toprule

\textbf{Method}& \textbf{Uncover} & \textbf{Cover} & \textbf{Curtain}  \\
\midrule
BC        & 85\% & 60\% & 60\%  \\
Retrieval & 90\% & 80\% & 80\%   \\
\bottomrule
\end{tabular}
\figvspace{}
\end{table}

\subsection{Deployment on Real Robot} 
To assess the real-world applicability of policies trained within MoDeSuite, we transfer learned models to physical hardware using the Boston Dynamics Spot robot. Specifically, we evaluate three representative tasks: Place, Drag, and Curtain. The first two tasks, which involve two types of elastic object manipulation, demonstrate promising sim-to-real transferability. In contrast, performance on the Curtain task reveals a noticeable sim-to-real gap, underscoring the challenges of visual domain generalization.

Figure~\ref{fig_drag_real} shows the physical experiment setup for the Drag task. We use a foam swimming noodle for the Place task, rubber stretching belts for Drag, and a 100cm × 120cm cloth for Curtain. State observations come from OptiTrack, and images are captured with an Intel RealSense D415. To evaluate the transferability of the algorithm trained in this task suite, we directly deploy two pre-trained agents per task in the real world without any fine-tuning.  

Table~\ref{tab:sim2real} presents a detailed comparison of performance between simulation and real-world evaluation for SAC and PPO on the Place and Drag tasks. Both methods demonstrate strong sim-to-real alignment in success rates across the two tasks, with the exception of SAC on the Drag task, where performance drops slightly in the real world. This discrepancy is primarily due to hardware limitations that prevent the execution of unsafe movements that SAC exploits in simulation. While the number of control steps differs due to the disparity in control frequencies (10 Hz in the real world versus 60 Hz in simulation), the overall task completion trends remain consistent across domains. These findings highlight the sim-to-real transferability of MoDeSuite with state-based observations.  Notably, PPO demonstrates not only high success rates but also consistent behavior across domains, as visualized in Figure~\ref{fig_drag_real}. The robot successfully completes the Drag task despite visible deviations in the deformable object’s trajectory, which we attribute to unavoidable differences in physical properties and real-world conditions. These results underscore the robustness of our approach and the transferability of learned behaviors in MoDeSuite, particularly when using state-based observations.

On the other hand, for the image-based state, we evaluated the policies trained in simulation on the Spot robot performing the curtain-opening task. While the policies are successful in simulation,  neither is able to complete the task in the real world. Specifically, the retrieval-based method managed to approach and make contact with the curtain in 2 out of 10 trials, whereas the behavior cloning (BC) policy failed to even reach the curtain. To investigate this discrepancy, we compared the observation trajectories and the encoded visual features from both domains. The observation trajectory recorded during manual teleoperation in the real world closely resembled the simulated one, indicating that the simulation captures the task dynamics with high fidelity. However, a t-SNE visualization of the encoded visual features revealed a clear separation between the simulation and real-world distributions. This suggests that the failure is primarily due to a visual domain gap, rather than a mismatch in task dynamics. These findings emphasize the need for stronger visual domain generalization and motivate future work in domain adaptation and representation learning for sim-to-real transfer in vision-based policies.

\section{Conclusion}
To address the gap in existing benchmarks for mobile deformable manipulation, we introduce the first comprehensive task suite, MoDeSuite, which includes both elastic deformable objects and plastic deformable fabrics. MoDeSuite includes five elastic tasks and three plastic tasks, supported by two types of robot configurations. The suite provides both state-based and image-based observation and offers controllers in joint space, task space, and hyper-mode. We evaluate two representative reinforcement learning algorithms and two imitation learning methods as baselines to facilitate further advancements in mobile deformable manipulation algorithms.

The performance of the trained agents highlights the significant challenges posed by mobile deformable manipulation, particularly due to the complex dynamics of the objects and the need for coordinated control across the robot body and arm. To evaluate the practical applicability of the proposed tasks, we directly deployed policies trained in simulation on real robotic platforms without any additional fine-tuning. The results highlight both the potential for sim-to-real transfer and the difficulty of achieving robust generalization in real-world settings. We believe that this benchmark provides a valuable testbed for systematically comparing mobile deformable manipulation approaches in simulation and will contribute to advancing the development of effective sim-to-real transfer techniques in this domain.

Looking ahead, we plan to extend this task suite by introducing additional elastic object shapes, such as toruses, to diversify the set of manipulable objects. While the size and shape of elastic objects can currently be adjusted via the Isaac Sim Graphical User Interface (GUI), we are considering programmatic methods in the future to enhance the flexibility of the suite.

\addtolength{\textheight}{-1cm}   



\bibliographystyle{IEEEtran}
\bibliography{tasksuiteref}

\end{document}